\documentclass{vgtc}                          

\graphicspath{{figures/}{pictures/}{images/}{./}} 

\usepackage{times}                     

\usepackage{tabu}                      
\usepackage{booktabs}                  
\usepackage{lipsum}                    
\usepackage{mwe}                       

\usepackage{mathptmx}                  
\usepackage{cuted}
\usepackage{caption}

\onlineid{0}

\vgtccategory{Research}

\vgtcinsertpkg

\author{
Nikita Kuzmin$^{\S}$\textsuperscript{*}
\thanks{e-mail: Nikita.Kuzmin@skoltech.ru} %
\and Yuhua Jin$^{\ddagger}$\textsuperscript{*}
\thanks{e-mail: yuhuajin@cuhk.edu.cn} %
\and Georgii Demianchuk$^{\S}$\textsuperscript{*}
\thanks{e-mail: Georgii.Demianchuk@skoltech.ru} %
\and Mariya Lezina$^{\S}$
\thanks{e-mail: Mariya.Lezina@skoltech.ru} %
\and Fawad Mehboob$^{\S}$
\thanks{e-mail: Fawad.Mehboob@skoltech.ru} %
\and Ivan Valuev$^{\S}$
\thanks{e-mail: Ivan.Valuev@skoltech.ru} %
\and Nikolai Lutsenko$^{\S}$
\thanks{e-mail: Nikolai.Lutsenko@skoltech.ru} %
\and Miguel Altamirano Cabrera$^{\S}$
\thanks{e-mail: M.Altamirano@skoltech.ru} %
\and Dzmitry Tsetserukou$^{\S}$
\thanks{e-mail: D.Tsetserukou@skoltech.ru}
}

\affiliation{
\scriptsize
$^{\S}$Intelligent Space Robotics Lab,
Skolkovo Institute of Science and Technology,
Moscow, Russian Federation \\

$^{\ddagger}$School of Science and Engineering,
The Chinese University of Hong Kong, Shenzhen,
Guangdong, China \\

\textsuperscript{*}These authors contributed equally to this work.
}

\title{OmniAI: A Surface-Adaptive Aerial Projection Interface for Human--Drone Interaction}

\teaser{
\vspace{-5mm}
  \includegraphics[width=1.1\textwidth,height=0.30\textheight,keepaspectratio]{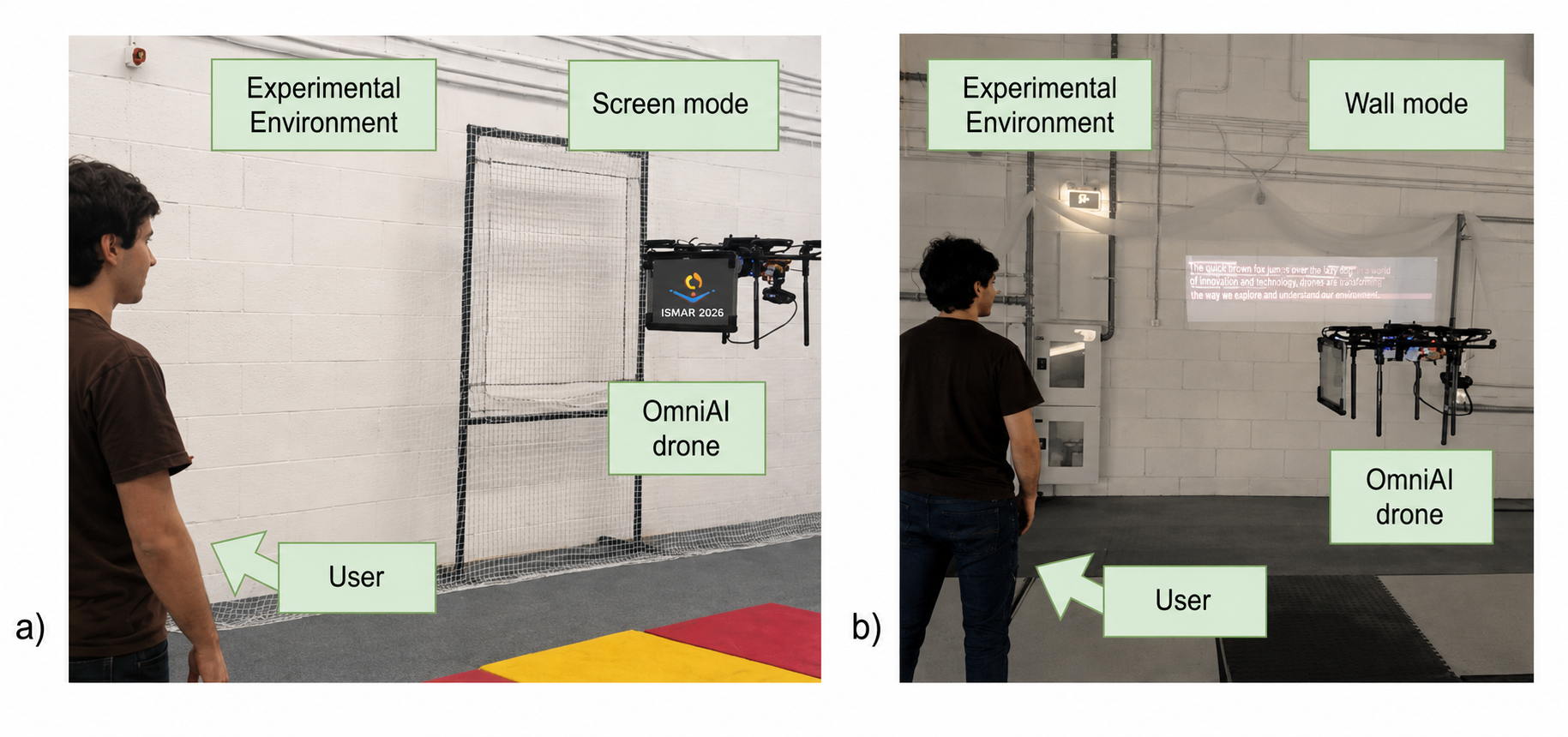}

    \vspace{-4mm}
  
    \caption{The OmniAI system: (a) onboard-screen mode for face-to-face dialogue; (b) wall-projection mode with gesture-based interaction.}
  }

\abstract{%
Drones in human environments often lack spatially grounded interfaces for situated communication. We present OmniAI, an embodied aerial agent that supports surface-adaptive interaction by switching projection between an onboard screen and nearby environmental surfaces. A servo-actuated MEMS laser projector renders text-and-image responses from a web-augmented LLM pipeline. Projection surfaces are detected online using RGB-D sensing and RANSAC plane fitting, without pre-mapped geometry. OmniAI provides functionally equivalent voice and gesture control for both drone motion and projected content. By combining speech, mid-air gestures, adaptive projection, and aerial mobility, OmniAI demonstrates a mobile spatial AR interface for context-aware human-drone interaction.
}

\keywords{Human-drone interaction, aerial interfaces, projection, spatial user interfaces, gesture interaction.}

\begin{document}


\firstsection{Introduction}

\maketitle

Drones are increasingly used as mobile interactive platforms, yet their communication bandwidth to nearby users remains constrained. HDI research has explored flight motion, gestures, touch, proxemics, emotion cues, and visual interfaces to make drones interpretable and controllable~\cite{cauchard2015drone, cauchard2016emotion, abtahi2017drone, cauchard2019drone}. Visual output is useful when a drone must present maps, instructions, retrieved images, or explanations in the user's physical environment. However, aerial visual interfaces face a practical tension: onboard displays remain available while the drone moves, whereas environmental projection provides a larger and more readable surface only when a suitable wall or screen exists nearby.

We investigate this tension through \emph{surface-adaptive aerial projection}: an interaction approach in which a drone selects between a small onboard display and a detected environmental surface at runtime. The contribution is not a new plane-fitting, speech-recognition, or gesture-recognition algorithm. Instead, the paper contributes a system-level interaction prototype that makes this switching behavior concrete and identifies constraints for spatial aerial interfaces.

OmniAI is a 1.2~kg prototype with a rotatable MEMS laser projector, onboard projection screen, RGB-D camera, ground-station perception pipeline, and gesture-based navigation of projected content. In wall-projection mode, the system detects a dominant planar surface, yaws the drone toward the plane, and then uses a pan--tilt mechanism for final projector alignment. Retrieved text and images are rendered as a multi-page booklet browsed with mid-air gestures.

This paper contributes: (1) a surface-adaptive aerial projection concept that treats onboard projection as fallback and wall projection as the preferred high-readability mode; (2) a working prototype and alignment pipeline for switching modes without pre-installed infrastructure; and (3) a preliminary evaluation of display surface, input modality, and component reliability, with explicit limitations.

\section{Related Work}

Human--drone interaction (HDI) has explored communication through gestures, touch, visual feedback, and expressive drone behavior. Foundational studies established natural and multimodal interaction paradigms, including \emph{Drone \& Me}, \emph{Drone Near Me}, and \emph{Drone.io}~\cite{cauchard2015drone, abtahi2017drone, cauchard2019drone}. Recent work has further investigated trustworthy and conversational aerial agents, highlighting the growing role of embodied AI in HDI~\cite{lingam2025human, jin2026hoverai}.

A complementary research direction investigates drones as aerial display platforms. \emph{BitDrones} demonstrated programmable aerial displays~\cite{gomes2016bitdrones}, while \emph{FlyMap} and \emph{LightAir} explored projection-based interaction using hovering drones~\cite{brock2018flymap,matrosov2016lightair}. These systems demonstrate the feasibility of aerial visualization but typically rely on a single display modality.

OmniAI builds upon these directions by investigating a surface-adaptive interaction paradigm rather than proposing new perception or projection algorithms. The system dynamically switches between an onboard display and wall projection according to the available interaction surface while presenting retrieval-augmented multimodal responses. Existing retrieval-augmented generation techniques are employed as enabling technologies~\cite{lewis2020retrieval}; the primary contribution lies in the interaction design that maintains communication continuity across private onboard and shared environmental displays.

\section{System}

\begin{figure*}[t]
\centering
\includegraphics[width=0.8\textwidth]{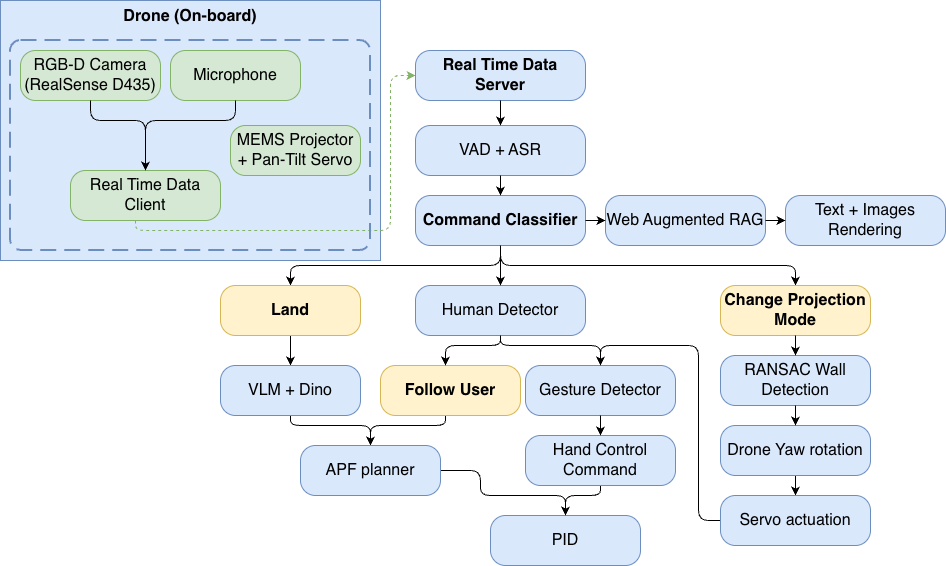}
\caption{System overview. OmniAI combines an aerial platform, steerable projection, RGB-D perception, speech input, and gesture interaction. The system switches visual output between an onboard screen and a detected environmental surface.}
\vspace{-3mm}
\label{fig:system}
\end{figure*}

OmniAI is a prototype aerial interface for spatially grounded human--drone interaction. Its central interaction mechanism is surface-adaptive projection: short feedback is shown on a small onboard screen, while longer or shared content is projected onto a nearby environmental surface. 
\subsection{Architecture}

\begin{figure}[t]
    \centering
    \includegraphics[width=0.8\columnwidth]{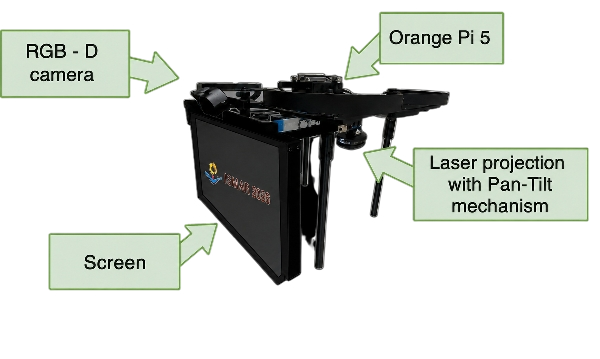}
    \vspace{-5mm}
    \caption{OmniAI hardware platform showing the Intel RealSense D435 RGB-D camera, Orange Pi~5 onboard computer, semi-rigid onboard projection screen, and MEMS laser projector mounted on a two-DOF servo pan--tilt mechanism.}
    \vspace{-5mm}
    \label{fig:hardware}
\end{figure}

The prototype uses a 1.2~kg drone equipped with an Orange Pi~5, an Intel RealSense D435 RGB-D camera, a close-talking microphone, a MEMS laser projector, and a two-DOF servo pan--tilt mount. A semi-rigid polycarbonate film mounted in front of the drone serves as the onboard projection surface. In screen mode, the projector faces this film; in wall mode, the pan--tilt mechanism redirects the projector toward an external surface. Figure~\ref{fig:hardware} shows the assembled platform.

OmniAI uses a distributed architecture. The drone streams audio, RGB, and depth data to a WiFi-connected ground-station PC. Computationally heavy modules run offboard, including VAD, ASR, language-model-based intent routing, surface detection, gesture recognition, object detection, retrieval, response generation, and rendering. The drone executes flight and projector commands received from the ground station. Thus, the current system is a distributed aerial interface prototype rather than a fully onboard autonomous agent.

The system processes two main input streams. The audio stream is used for spoken commands and open-ended queries. The RGB-D stream is used for wall detection, gesture tracking, object localization, and projection-aware scene understanding. Speech and gestures are mapped to a shared command interface, allowing users to control both projected content and drone behavior.

\subsection{Display Modes and Surface Selection}

OmniAI supports two display modes. In \emph{screen mode}, the projector displays content on the onboard polycarbonate film. This mode is used for short answers, status feedback, confirmations, and fallback interaction when no suitable environmental surface is available.

In \emph{wall mode}, the system searches for a planar projection surface using the RealSense depth frame. The depth image is unprojected into a 3D point cloud using the camera intrinsics. Points outside the operational distance range are removed, and RANSAC plane fitting is applied to identify a dominant planar surface. The selected plane is refined using its inlier points, and the estimated normal is used to determine the required drone yaw and projector orientation.

The alignment procedure is sequential. First, the drone yaws toward the selected surface normal. After stabilization, the pan--tilt mount adjusts the projector direction. This yaw-then-pan--tilt sequence reduces visible image drift during mode switching and improves projection readability.

For scene-level perception, OmniAI also uses a vision pipeline combining text-conditioned object detection and vision-language interpretation. Grounding DINO localizes objects from natural-language descriptions, while a vision-language model helps disambiguate targets and interpret scene context. This perception layer is used for projection-aware target selection and for converting visual targets into control commands when the user asks the drone to move relative to objects or surfaces.

\subsection{Interaction Flow}

The interface is organized around five user-facing modes: listening, onboard display, surface search, wall projection, and browsing. Users do not directly manipulate the internal state machine. Instead, speech and gestures are interpreted through a shared command layer that maps inputs to projection switching, content navigation, information requests, or drone motion.

For speech interaction, audio is first processed by Voice Activity Detection (VAD) to isolate speech segments. The detected segments are transcribed by an Automatic Speech Recognition (ASR) module and passed to a language-model-based intent router. The router classifies each utterance as either an executable command or an information-seeking query. Motion commands, such as following, landing, directional movement, or yaw rotation, are sent to the flight-control module. Projection commands trigger surface detection and servo actuation. Gesture-mode commands enable or disable hand tracking. Query intents are forwarded to the retrieval-augmented generation pipeline.

Gesture interaction uses the RGB-D stream from the onboard camera. Hand landmarks are detected from the RGB image and combined with depth information when spatial control is required. A lightweight rule-based controller maps recognized gesture states to high-level actions such as page navigation, projection-mode changes, or drone-control commands. Speech and gestures therefore support overlapping functions: both can control content and drone behavior, while gestures provide a spatial, touchless alternative that is especially useful during wall projection.

\subsection{Content Retrieval and Rendering}

When the intent router identifies an information-seeking query, OmniAI forwards the transcript to a retrieval-augmented generation pipeline. The system first searches a local vector store of curated facts. If local retrieval confidence is insufficient, it augments the context with web search results before generation. The language model produces a structured response consisting of a text answer and, when relevant, an image-search query.

The rendering module adapts the generated response to the active display mode. In onboard-display mode, answers are shortened and rendered as compact text because the available projection area is limited. In wall-projection mode, longer responses are formatted as a multi-page layout combining text and retrieved images. Users can browse these pages using either speech or gestures. Thus, OmniAI adapts both the output surface and the visual structure of the response.

\subsection{Scope}

The current system targets indoor environments with flat projection surfaces such as walls and whiteboards. It does not yet handle curved surfaces, strong sunlight, highly reflective materials, or dense occlusions. The drone also depends on an offboard computer for heavy perception and language processing. These constraints limit deployment, but they clarify the research scope: OmniAI is a prototype for studying surface-adaptive aerial interfaces and interaction transitions between onboard and environmental displays.

\section{Preliminary Evaluation}

We evaluated OmniAI as a prototype of surface-adaptive aerial projection rather than as a definitive comparison of all possible drone communication modalities. This distinction is important: the current study characterizes feasibility, workload trends, and component reliability, but it does not isolate dynamic surface switching against a fixed-mode baseline. We therefore report the results as preliminary evidence for the design space and avoid claims of general superiority.

\begin{table*}[t]
\centering
\caption{NASA-TLX ratings (mean $\pm$ SD, scale 1--20). Lower values indicate lower perceived workload. The study is exploratory with \(N=10\) participants.}
\vspace{-2mm}
\label{tab:nasatlx}

\footnotesize
\setlength{\tabcolsep}{4.5pt}
\renewcommand{\arraystretch}{1.08}

\resizebox{0.9\textwidth}{!}{%
\begin{tabular}{lcccccccc}
\toprule
& \multicolumn{2}{c}{Noise}
& \multicolumn{2}{c}{Surface}
& \multicolumn{2}{c}{Content}
& \multicolumn{2}{c}{Input} \\

\cmidrule(lr){2-3}
\cmidrule(lr){4-5}
\cmidrule(lr){6-7}
\cmidrule(lr){8-9}

Subscale
& Noise & No noise
& Screen & Wall
& Text & Text+Image
& Speech & Gesture \\

\midrule
Mental Demand   & $9.8\pm4.2$ & $2.8\pm1.5$ & $9.4\pm7.0$ & $3.0\pm2.5$ & $9.4\pm7.0$ & $4.6\pm3.1$ & $7.4\pm3.6$ & $3.4\pm2.3$ \\
Physical Demand & $6.8\pm3.7$ & $2.6\pm1.0$ & $7.8\pm6.8$ & $1.8\pm1.8$ & $7.8\pm6.8$ & $3.8\pm4.5$ & $6.8\pm3.7$ & $2.6\pm1.5$ \\
Performance     & $5.6\pm2.2$ & $2.4\pm2.1$ & $9.4\pm7.0$ & $2.8\pm2.6$ & $8.0\pm6.9$ & $3.4\pm4.4$ & $7.4\pm2.2$ & $2.0\pm1.4$ \\
Effort          & $4.2\pm1.9$ & $2.2\pm1.1$ & $9.8\pm8.0$ & $2.8\pm4.0$ & $9.6\pm8.0$ & $3.6\pm3.4$ & $5.0\pm2.3$ & $2.4\pm0.5$ \\
Frustration     & $8.8\pm4.9$ & $2.5\pm1.5$ & $10.0\pm7.2$ & $4.4\pm3.4$ & $10.0\pm7.2$ & $6.0\pm5.5$ & $4.4\pm3.4$ & $3.0\pm1.8$ \\
\bottomrule
\end{tabular}%
}
\vspace{-3mm}
\end{table*}

Experiments were conducted in a controlled indoor laboratory of approximately $6\,\mathrm{m} \times 6\,\mathrm{m}$ under moderate ambient lighting. Ten participants completed interaction sessions involving projection mode changes, wall-projected content browsing, speech commands, and gesture-based navigation. The study focused on three questions: whether wall projection improves perceived workload relative to the onboard screen, whether gesture input remains usable under drone noise, and whether the prototype components are sufficiently reliable for controlled demonstrations.
\subsection{Component Reliability}

We evaluated the prototype in a controlled indoor laboratory setting. These measurements are not intended as algorithmic benchmarks; they characterize whether the implemented system was reliable enough to support short surface-adaptive interaction sessions.

We first tested the vision pipeline used for object-aware interaction. Table~\ref{tab:detection_summary} reports category-level detection rate, mean confidence, and maximum successful detection distance for everyday objects used in the demonstrations.

\begin{table}[t]
\centering
\caption{Object-level detection results for the VLM + Grounding DINO perception pipeline. Detection denotes successful recognition within the tested trials; confidence is averaged over successful detections.}
\vspace{-2mm}
\label{tab:detection_summary}
\scriptsize
\setlength{\tabcolsep}{3.5pt}
\renewcommand{\arraystretch}{1.05}
\resizebox{0.75\columnwidth}{!}{%
\begin{tabular}{lccc}
\toprule
Object & Detection & Confidence & Max dist. \\
       & (\%) & (\%) & (m) \\
\midrule
Ball       & 75  & 65.7 & 2.13 \\
Bottle     & 100 & 76.2 & 3.60 \\
Chair      & 100 & 80.7 & 1.28 \\
Cube       & 65  & 68.6 & 4.05 \\
Cup        & 100 & 73.7 & 3.88 \\
Headphones & 100 & 84.8 & 4.16 \\
Person     & 100 & 64.5 & 3.67 \\
Plant      & 60  & 60.1 & 1.37 \\
\bottomrule
\end{tabular}
}
\vspace{-4mm}
\end{table}

Object recognition varied by category. Larger or visually distinctive objects such as bottles, cups, headphones, chairs, and persons were detected reliably in the tested scenes, whereas smaller or less distinctive categories such as plants, cubes, and balls were less stable. This limitation affects commands that depend on visual target selection, but not projection-mode switching or content browsing.

For flight-related behavior, static obstacle avoidance succeeded in all 100 trials, with mean path efficiency of 0.83 and mean speed of 0.97~m/s. Dynamic obstacle avoidance, tested with a person crossing the drone's path, succeeded in 84\% of 100 trials, with mean path efficiency of 0.91 and mean speed of 1.35~m/s. These results indicate that the prototype can support short indoor interaction sessions, although dynamic human motion remains a reliability bottleneck.

Input reliability also differed across modalities. Gesture commands achieved 58.8\% success over 80 command trials, while voice commands achieved 46.3\% success over the same number of trials. This gap is consistent with the acoustic limitations of the aerial platform: propeller noise degraded speech intelligibility despite the VAD and ASR pipeline. Consequently, speech is most appropriate for discrete high-level commands, such as mode switching or query input, while gestures are better suited for repeated content navigation and spatial control during wall projection.
\subsection{User Study}

We conducted an exploratory user study with (N=10) participants using a modified NASA-TLX questionnaire. Participants evaluated five workload subscales across four interface factors: drone noise, projection surface, content type, and interaction modality. Table~\ref{tab:nasatlx} reports the mean and standard deviation for each condition.


Because of the limited sample size, we treat the statistical analysis as exploratory. After Holm correction, the clearest significant result was that speech interaction was rated as more mentally demanding than gesture interaction during drone operation (speech: (M=7.40), gesture: (M=3.40), (p=0.0289)). Descriptively, wall projection produced lower workload ratings than onboard-screen projection across all five subscales, but these differences should be interpreted as preliminary.

These findings support a modest conclusion: wall projection appears preferable when a suitable surface is available, while the onboard screen is better understood as a close-range or fallback channel. The study does not prove that adaptive switching itself improves task performance; a future evaluation should compare adaptive switching against fixed onboard-only and wall-only baselines.

\section{Limitations}

Flight time is approximately 12 minutes due to the added projector, screen, camera, and compute payload. The MEMS projector is suitable only for controlled indoor lighting; bright environments reduce legibility. Gesture recognition assumes a single active user and remains sensitive to viewpoint, hand pose, and partial occlusion. The current method does not include documented ego-motion compensation, so we avoid claiming stabilization beyond relative hand-landmark normalization. RANSAC plane fitting fails on curved or heavily occluded surfaces and assumes a visible dominant plane. Most computation runs on a WiFi-connected ground station, limiting range and robustness. Public deployment would also require visible sensing indicators, consent mechanisms, data minimization, and open-licensed image sources.

\section{Conclusion}

OmniAI explores surface-adaptive aerial projection as a practical HDI and spatial-interface concept. The prototype demonstrates switching between an onboard display and a detected wall, projection alignment through a yaw-then-pan--tilt sequence, and retrieval-based text--image content as a gesture-navigable booklet. The evaluation supports modest conclusions: wall projection is preferable for readable content when available, gesture interaction is more usable than speech under drone noise, and current reliability supports controlled indoor demonstrations but not robust deployment. The next step is a focused dynamic-switching study against fixed-display baselines.

\section*{Acknowledgements} 
Research reported in this publication was financially supported by the RSF grant No. 24-41-02039.

\end{document}